\documentclass[runningheads]{llncs}
\usepackage{graphicx}

\usepackage{subcaption}
\usepackage{amsmath}
\usepackage{algorithm}
\usepackage{algpseudocode}
\usepackage{hyperref}
\usepackage{url}
\usepackage{float}
\usepackage{multirow}
\usepackage{graphicx} 

\usepackage[utf8]{inputenc} 
\usepackage[T1]{fontenc}    
\usepackage{hyperref}       
\usepackage{url}            
\usepackage{booktabs}       
\usepackage{amsfonts}       
\usepackage{nicefrac}       
\usepackage{microtype}      
\usepackage{xcolor}         
\usepackage{amsmath}
\usepackage{tikz}
\usepackage{framed}
\usepackage{multirow}
\usepackage{caption}
\usepackage{titlesec}
\usepackage{pgfplots}

\usepackage{tikz}
\usetikzlibrary{positioning, shapes, calc, plotmarks}

\begin{document}
\title{Visual Concept Networks: A Graph-Based Approach to Detecting Anomalous Data in Deep Neural Networks\thanks{This research was supported in part by NSF awards \#2104377, \#2112606 and \#2117439.}}
\titlerunning{Visual Concept Networks}
%
\author{Debargha Ganguly\inst{1} \and
Debayan Gupta\inst{2} \and
Vipin Chaudhary\inst{1}}
\authorrunning{D. Ganguly et al.}
%
\institute{Case Western Reserve University \\
\email{\{debargha.ganguly \& vipin\}@case.edu}
\and
Ashoka University\\
\email{\{debayan.gupta\}@ashoka.edu.in}}
\maketitle              
\begin{abstract}

Deep neural networks (DNNs), while increasingly deployed in many applications, struggle with robustness against anomalous and out-of-distribution (OOD) data. Current OOD benchmarks often oversimplify, focusing on single-object tasks and not fully representing complex real-world anomalies. This paper introduces a new, straightforward method employing graph structures and topological features to effectively detect both far-OOD and near-OOD data. We convert images into networks of interconnected human understandable features or visual concepts. Through extensive testing on two novel tasks, including ablation studies with large vocabularies and diverse tasks, we demonstrate the method's effectiveness. This approach enhances DNN resilience to OOD data and promises improved performance in various applications.

\keywords{Out-of-Distribution Detection  \and Deep Neural Networks Robustness \and Graph-Based Anomaly Detection}

\end{abstract}

\section{Introduction}

Trustworthy machine learning (ML) systems need to refer low-confidence decisions to human experts. They often rely on the closed-world assumption, expecting test data to mirror the training data's distribution \cite{krizhevsky2012imagenet}. However, real-world scenarios frequently encounter deviations from this ideal, particularly in open-world settings where test samples might be out-of-distribution (OOD) \cite{drummond2006open}. These OOD samples arise from either semantic shifts (different classes) \cite{hendrycks2016baseline} or covariate shifts (different domains) \cite{ben2010theory}, necessitating specialized handling. Related areas such as outlier, anomaly, novelty detection, and open set recognition \cite{wang2019progress} \cite{ruff2021unifying} \cite{miljkovic2010review} \cite{mahdavi2021survey} share techniques for managing such data variances.

\textbf{Issues: }The literature often overlooks significant challenges in validating OOD detection methods. A key issue is the unrealistic perfection suggested by many studies, where the area under the receiver operating characteristic curve (AUROC) scores near or reach 100 \cite{cao2022deep}. This raises a critical question: do these tasks accurately represent the complexities of OOD detection, or are they overly simplistic? The gap between academic datasets and real-world applications highlights the complexity of the issue. Using simpler datasets like CIFAR-10 versus CIFAR-100, or comparing MNIST to SVHN, oversimplifies the OOD detection challenge. These 'toy tasks' don't fully represent the range and intricacy of data encountered in actual production environments \cite{hendrycks2019scaling}. To solve this the Open OOD benchmark was introduced, with more difficult and nuanced tasks \cite{yang2022openood}. However, performance on these benchmarks are also saturated at near perfect levels \cite{park2023nearest}.

\textbf{Benchmark contributions} To address existing gaps, we introduce two new OOD detection tasks using the LSUN and ImageNet datasets for more accurate in-distribution representation. These machine-annotated datasets, with their diverse and extensive distribution, aim to overcome the limitations of current OOD benchmarks. This enables researchers and practitioners to engage with a new benchmark that is both realistic and challenging, enhancing OOD detection methods. Additionally, because of the multi-object complex scenes, the OOD detection method must also have some level of explainability for "why" certain data points are classified as OOD, building upon tasks such as the anomaly segmentation task in \cite{hendrycks2019scaling}.

\textbf{Algorithmic Intuition} Our approach is rooted in the idea that graphs, with their ability to capture complex relationships, are ideal for analyzing entities and interactions in intricate domains. We hypothesize that using graph-based representations of visual features, which are interpretable by humans, will more effectively encode domain knowledge and identify OOD scenarios. This is because it allows the AI system to utilize the intricate relationships among visual concepts and infer latent semantics, while maintaining explanations that are understandable to humans.

\section {Background and Related Work}

Detecting out-of-distribution (OOD) points in lower-dimensional spaces, as discussed in Pimentel et al.'s review \cite{pimentel2014review}, typically relies on techniques such as density estimation, nearest neighbor algorithms, and clustering, which predict OOD points by assessing their density or distance from cluster\cite{theis2015note}.

\textbf{Techniques: }Research in OOD data is concentrated on network adjustments and the development of specialized scoring functions. Methods like network truncation, including techniques such as ODIN\cite{liang2017enhancing}, ReAct \cite{sun2021react}, and DICE \cite{sun2022dice}, modify a network's signals or weights to differentiate regular data from OOD data, but they require a complementary scoring function for effective OOD detection. These scoring functions can generally be surrogate classifier-based, like MSP\cite{hendrycks2016baseline} and the energy function \cite{liu2020energy}, which rely on the network's classification layer, or distance-based, such as the Mahalanobis detector \cite{lee2018simple} and KNN\cite{sun2022out}, which assess the dissimilarity of input from regular data based on features. Apart from these, post-hoc methods like temperature scaling, gradient input modifications, and statistical measurements have been introduced, offering the advantage of being easily integrated into various model architectures without requiring major alterations to the network itself. 

Additionally, there's a focus on enhancing OOD detection through training-time regularization techniques that aim to refine model training. This includes incorporating confidence estimation branches \cite{liang2017enhancing}, altering loss functions\cite{hendrycks2016baseline}, or utilizing contrast learning objectives \cite{wang2020understanding} to develop stronger models with more precise uncertainty estimates, albeit at the cost of increased computational resources. Some strategies also involve the integration of external OOD samples during the training phase to sharpen the model's ability to discern between in-distribution and OOD data \cite{hendrycks2018deep}. These methods range from promoting varied predictions on OOD samples \cite{jeong2020ood} to employing clustering techniques to sift out in-distribution samples. Although leveraging external data is a prevalent practice, particularly in the industry, it presents challenges such as selecting suitable data sets and preventing the model from overfitting to specific OOD samples.

Recent advancements in out-of-distribution (OOD) detection primarily utilize deep neural networks (DNNs). Techniques like convolutional neural networks (CNNs) for anomaly detection \cite{sabokrou2018deep}, and methods combining transfer and representation learning \cite{andrews2016transfer}\cite{fort2021exploring}, are prevalent. In critical areas like healthcare, generative adversarial networks (GANs) are used for unsupervised tasks \cite{schlegl2017unsupervised}. However, these approaches, involving additional neural layers or alterations, face a notable risk: DNNs can be overly confident in incorrect decisions in OOD contexts \cite{hendrycks2016baseline} \cite{lakshminarayanan2017simple}\cite{guo2017calibration}.

Before the advent of deep learning, image analysis techniques like Visual Bag-of-Words (BoW) treated image regions as 'words' to create a visual vocabulary and represent images as histograms, whereas Probabilistic Latent Semantic Analysis (pLSA) interpreted images as compositions of latent visual topics, considering each 'visual word' as a manifestation of these topics. CSKGs are emerging as valuable sources of domain-specific knowledge, aiding in tasks like question answering and planning \cite{ilievski2021cskg}\cite{guan2019knowledge}. In our study, we leverage these graphs, combined with the latest in geometric learning, to enhance OOD data detection.

\section{Problem Setup}

This paper addresses the challenge of distinguishing in-distribution (ID) and out-of-distribution (OOD) images using a pre-trained neural network. We define two distributions within the feature space $\mathcal{X}$: $D_{in}$ for ID and $D_{out}$ for OOD. The ID dataset $\mathcal{D}^{in}$ is composed of pairs $\left(\textbf{x}^{in}, y^{in}\right)$, where $\textbf{x}$ is the input feature and $y^{in}$ belongs to the set of class labels $\mathcal{Y}^{in} := {1, \ldots, K}$. The OOD dataset $\mathcal{D}^{out}$ includes pairs $\left(\textbf{x}^{out}, y^{out}\right)$, with $y^{out}$ in $\mathcal{Y}^{out} := {K+1, \ldots, K+O}$, ensuring $\mathcal{Y}^{out} \cap \mathcal{Y}^{in} = \emptyset$.

Our technique involves representing each input image $X$ as a network of visual features, with a high-dimensional embedding in latent space $Z$, obtained from a mixture distribution $P_{X*Z}$. The latent representation $Z$ is then analyzed to determine whether $X$ originates from the in-distribution $\mathcal{D}{in}$ or not. We explore two formulations, one with access to OOD samples during the training phase, similar to \cite{hendrycks2018deep} testing performance against a held-out set, and a the other being zero-shot with no access to OOD outputs.

\textbf{Types of Shift: }Model performance is subject to different distribution shifts: \textit{covariate shift} impacts the input space $\mathcal{X}$, while \textit{semantic shift} affects the label space $\mathcal{Y}$. With a joint distribution $P(X, Y)$ over $\mathcal{X} \times \mathcal{Y}$, shifts can alter either the marginal $P(X)$ or both $P(X)$ and $P(Y)$. Notably, changes in $P(Y)$ inherently influence $P(X)$.

Covariate shifts, seen in cases like adversarial attacks \cite{goodfellow2014explaining}, domain changes \cite{quinonero2008dataset}, and style variations \cite{gatys2016image}, mainly impact $P(X)$, testing models' generalization while keeping $Y$ constant. Semantic shifts, however, involve changes in the label space $Y$ between in-distribution (ID) and out-of-distribution (OOD) data, crucial in many detection tasks. Models must be cautious in making predictions here. OOD detection's effectiveness largely depends on the semantic similarity between outliers and inliers. Near OOD tasks, like shifts from SVHN to MNIST, pose greater challenges, with current methods achieving about 93\% AUROC \cite{fort2021exploring}. Far OOD tasks, with clearer semantic differences, generally see AUROCs near 99\% \cite{fort2021exploring}.

\begin{table*}[htb]
\centering
\caption{OOD detection tasks based on \cite{yu15lsun}}
\label{task-table}
\setlength{\tabcolsep}{0.8em} 
{\renewcommand{\arraystretch}{1.2}
\begin{tabular}{p{5cm}p{6cm}}
\toprule
\multicolumn{1}{c}{\textbf{Far-OOD Tasks}} & \multicolumn{1}{c}{\textbf{Near-OOD Tasks}} \\
\midrule
Bridge vs (Classroom, Conf. Room, Dining, Kitchen, Living Room, Restaurant, Bedroom) & Bedroom vs (Classroom, Conf. Room, Dining, Kitchen, Living Room, Restaurant); \\
& Living Room vs Restaurant \\
\addlinespace
\textbf{Church Outdoor} vs (Classroom, Conf. Room, \textbf{Dining}, Kitchen, Living Room, Restaurant, Bedroom) & Church Outdoor vs (Tower, Bridge), Bridge vs Tower; \\
& Classroom vs (Conf. Room, Dining, Kitchen, Living Room, Restaurant); \\
& Kitchen vs (Living Room, Restaurant) \\
\addlinespace
Tower vs (Classroom, Conf. Room, Dining, Kitchen, Living Room, Restaurant, Bedroom) & Conf. Room vs (Dining, Kitchen, Living Room, Restaurant); \\
& \textbf{Dining} vs (Kitchen, Restaurant, \textbf{Living Room}) \\
\bottomrule
\end{tabular}}
\end{table*}

\subsection{Benchmarking Tasks}

Our benchmarking innovation uses the Large-scale Scene Understanding (LSUN) dataset \cite{yu15lsun} to enrich out-of-distribution (OOD) detection benchmarks. LSUN's diverse range of real-world images, from 120k to 3 million across ten categories like bedrooms and living rooms, forms the basis of our benchmarks. We classify these categories into discrete domains and create pairwise combinations, distinguishing them as far-OOD or near-OOD based on semantic closeness. This setup allows for a more nuanced and rigorous evaluation of OOD detection capabilities.

Further, we incorporate the ImageNet dataset, structured on the WordNet lexical hierarchy, to introduce a complex task. We focus on its broad top-tier classes like 'entities' and 'natural objects', each encompassing varied subclasses. This approach tests models' ability to classify images and identify their broader distributions, simulating real-world applications where understanding both specific details and overarching contexts is crucial.

Our benchmarks mark a significant step forward in OOD detection research. Unlike foundational datasets like CIFAR-10 and CIFAR-100, which focus on single-object images, our approach reflects the complexity and diversity of real-world visual data, often featuring multiple objects across classes. This design ensures models are tested against realistic, multifaceted scenarios, elevating the standard for OOD detection task evaluations.

\begin{figure*}[htb]
    \centering
    \resizebox{\textwidth}{!}{

    \begin{tikzpicture}

            \node[draw, rectangle, minimum width=4cm, minimum height=3cm, fill=gray!30, label=below:{Images \& their visual features}] (image) {};

\foreach \i in {1,2,3} {
    \node[draw, rectangle, minimum width=4cm, minimum height=3cm, fill=gray!20, shift={(-0.05*\i cm,0.05*\i cm)}] at (image) {};
}


\draw[red, thick] ([shift={(0.5cm,0.5cm)}]image.south west) rectangle ([shift={(1.5cm,1.5cm)}]image.south west);
\draw[red, thick] ([shift={(2.7cm,1.7cm)}]image.south west) rectangle ([shift={(1.2cm,2.2cm)}]image.south west);
\draw[red, thick] ([shift={(2cm,0.3cm)}]image.south west) rectangle ([shift={(0.8cm,1.8cm)}]image.south west);
\draw[red, thick] ([shift={(0.2cm,2.2cm)}]image.south west) rectangle ([shift={(1.6cm,2.8cm)}]image.south west);
\draw[red, thick] ([shift={(3.2cm,0.8cm)}]image.south west) rectangle ([shift={(2cm,1.4cm)}]image.south west);
\draw[red, thick] ([shift={(0.7cm,2.5cm)}]image.south west) rectangle ([shift={(1.4cm,2.9cm)}]image.south west);
\draw[red, thick] ([shift={(2.8cm,0.5cm)}]image.south west) rectangle ([shift={(2.2cm,1.2cm)}]image.south west);
\draw[red, thick] ([shift={(0.4cm,0.8cm)}]image.south west) rectangle ([shift={(0.9cm,1.3cm)}]image.south west);

\draw[dashed] (3.5cm, 2.5cm) -- (3.5cm, -2.5cm);

\begin{scope}[scale=0.4, transform shape, xshift=10cm]

    \foreach \i in {1,2,3} {
    \node[draw, rectangle, minimum width=10cm, minimum height=12cm, fill=gray!20, right=1.5cm of image, shift={(10cm-0.15*\i cm,-0.05*\i cm)}] at (image) {};
}

    \node[draw, circle, fill=green!20, minimum size=1cm, right=10cm of image] (node1) {};
    \node[draw, circle, fill=green!20, minimum size=0.8cm, above right=1.4cm and 3cm of node1] (node2) {};
    \node[draw, circle, fill=green!20, minimum size=1.2cm, below=2cm of node1] (node3) {};
    \node[draw, circle, fill=green!20, minimum size=1.5cm, above=1.5cm of node1] (node4) {};
    \node[draw, circle, fill=green!20, minimum size=1cm, below right=1cm and 2.5cm of node3] (node5) {};
    \node[draw, circle, fill=green!20, minimum size=1.3cm, above left=1cm and 1.5cm of node2] (node6) {};
    \node[draw, circle, fill=green!20, minimum size=0.9cm, below left=1.4cm and 2cm of node3] (node7) {};
    \node[draw, circle, fill=green!20, minimum size=0.7cm, left=2.5cm of node1] (node8) {};
    \node[draw, circle, fill=green!20, minimum size=0.5cm, left=1.5cm of node1] (node9) {};
    \node[draw, circle, fill=green!20, minimum size=1.2cm,  above left=1cm and 1.5cm of node1] (node10) {};

    \foreach \i in {1,...,10} {
        \foreach \j in {1,...,10} {
            \pgfmathparse{rand}
            \ifdim\pgfmathresult pt > 0.2pt 
                \ifnum\i<\j 
                    \draw (node\i) -- (node\j);
                \fi
            \fi
        }
    }
\end{scope}

\node[anchor=north west, font=\small] at (4.5cm, 2.25cm) {Visual Feature Networks};

\draw[dashed] (9.5cm, 2.5cm) -- (9.5cm, -2.5cm);

\begin{scope}[xshift=12cm] 
    \node[anchor=south west, font=\small] at (-2,-1.8) {Whole Graph};
    \node[anchor=south west, font=\small] at (-2,-2.1) {Embeddings};
    \draw[thick, ->] (0,-2.5) -- (0,2);
    \draw[thick, ->] (-1.5,0) -- (2,0);

    \def\RedCenterX{0.5}
    \def\RedCenterY{0.5}
    \def\RedVariance{0.3}

    \def\BlueCenterX{-0.5}
    \def\BlueCenterY{-0.5}
    \def\BlueVariance{0.7}

    \def\GreenCenterX{0.5}
    \def\GreenCenterY{-0.5}
    \def\GreenVariance{0.5}
    
    \foreach \i in {1,...,60} {
        \path[draw, fill=red] (\RedCenterX + rand*\RedVariance, \RedCenterY + rand*\RedVariance) circle (1.5pt);
    }

    \foreach \i in {1,...,40} {
        \path[draw, fill=blue] (\BlueCenterX + rand*\BlueVariance, \BlueCenterY + rand*\BlueVariance) circle (1.5pt);
    }

    \foreach \i in {1,...,20} {
        \path[draw, fill=green] (\GreenCenterX + rand*\GreenVariance, \GreenCenterY + rand*\GreenVariance) circle (1.5pt);
    }
\end{scope}

    \end{tikzpicture}
    }
\caption{An illustrative representation of the image-to-graph transformation process for out-of-distribution detection. The diagram commences with input images and their associated visual features, demarcated by bounding boxes. This visual information is then channeled into a graph structure, with nodes demonstrating unique visual elements and edges establishing interconnections. The concluding stage showcases the embedding of the entire graph into a 2D space, where similar visual patterns manifest as proximal clusters.}
    \label{fig:enter-label}
\end{figure*}
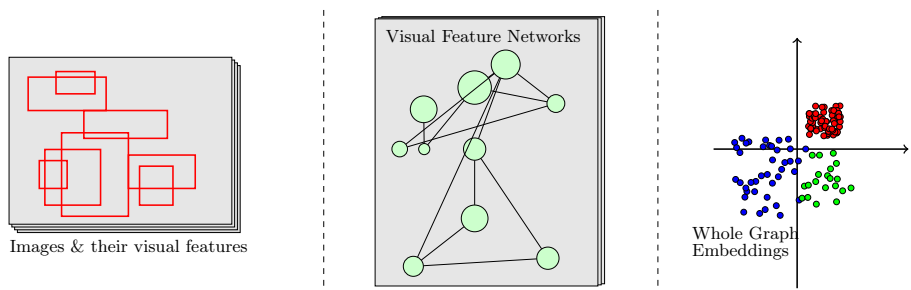

\subsection{Graph Embedding Algorithms}
In our approach, each unique concept from the visual vocabulary is represented as a node in the graph. Edges are constructed only between nodes present in the image. Given multiple object pairs, multiple edges are drawn. The weight of each edge is determined based on both the Intersection over Union (IoU) score, denoted by the Jaccard Index \( J \), and the euclidean distance between the centroids of the bounding boxes. Formally, for a graph \( G(x) = (V, E) \) where \( G(x) \in G \), the weight of the edge between objects \( \text{obj}_1 \) and \( \text{obj}_2 \) is defined as:

\[
E_{\text{weight}}^{\text{obj}_1,\text{obj}_2} = 
\begin{cases} 
1 + \lVert \text{obj}_1, \text{obj}_2 \rVert \times J(\text{obj}_1, \text{obj}_2) & \text{if weighted} \\
1 & \text{if unweighted}
\end{cases}
\]

\begin{enumerate}
\item \textbf{Graph2Vec} \cite{narayanan2017graph2vec} uses Weisfeiler-Lehman tree features to generate a 128-dimensional embedding with feature co-occurrence matrices. Parameters: 2 Weisfeiler-Lehman iterations, 10 epochs, learning rate of 0.025.
\item \textbf{Wavelet Characteristic} \cite{wang2021graph} employs wavelet function weights and node features to produce a 1000-dimensional embedding. Parameters: $\tau$= 1.0, $\theta_{max}$=2.5, 5 function evaluations.
\item \textbf{LDP} \cite{cai2018simple} calculates degree profile histograms to form a 160-dimensional graph representation. Parameters: 32 histogram bins.
\item \textbf{Feather Graph} \cite{rozemberczki2020characteristic} uses random walk weights for node features to create a 500-dimensional embedding. Parameters: $\theta_{max}$=2.5, 25 evaluation points.
\item \textbf{GL2Vec} \cite{chen2019gl2vec} leverages line graphs and edge features, similar to Graph2Vec, to form a 128-dimensional embedding.
\item \textbf{NetLSD} \cite{tsitsulin2018netlsd} uses the heat kernel trace for a 250-dimensional embedding. Parameters: 200 eigenvalue approximations, 250-time scale steps.
\item \textbf{SF} \cite{de2018simple} derives a 128-dimensional embedding from the lowest eigenvalues of the normalised laplacian.
\item \textbf{FGSD} \cite{verma2017hunt} creates a 200-dimensional embedding using spectral features of the normalized laplacian. Parameters: 200 histogram bins.
\end{enumerate}

Despite considering Invariant Graph Embeddings (IGE) \cite{galland2019invariant} and GeoScattering, challenges arose. IGE faced infinite path length issues, and GeoScattering necessitated fully connected graphs, which our dataset didn't satisfy.

\subsection{Evaluation Metrics}
To assess our system's effectiveness in distinguishing in-distribution and out-of-distribution samples, consistent with OOD detection literature \cite{fort2021exploring,liang2017enhancing}, we use several metrics, presenting weighted averages for comprehensive understanding. The Area Under the Receiver Operating Characteristic curve (AUROC) \cite{davis2006relationship}, a threshold-independent metric, measures the trade-off between true positive rate (TPR) and false positive rate (FPR), indicating a perfect detector at 100\%. The Area Under the Precision-Recall curve (AUPR) \cite{manning1999foundations,saito2015precision}, also threshold-independent, evaluates the precision-recall graph, especially useful in scenarios with infrequent anomalies. It reports the average precision (AP) score, summarizing the precision-recall curve. Lastly, the F1 Score, the harmonic mean of precision and recall, is crucial in OOD detection for balancing false positives and negatives, providing a single metric to gauge this balance. For multi-class classification in identifying originating distributions, we use the One-vs-Rest (OVR) approach for AUROC.

\subsection{Results \& Discussion} \label{results}
In this section, we first demonstrate our method's effectiveness in detecting OOD datapoints on the LSUN-based benchmark (in table 1). We employ the DETIC model \cite{zhou2022detecting}, an advanced open vocabulary object detection system developed by Meta, as our primary tool for feature extraction. Renowned for its comprehensive coverage and versatility, DETIC can discern and categorize a wide array of object classes, making it particularly apt for our assessment needs on diverse benchmarks.

Due to computational constraints, we test 100,000 images in each in-distribution and out-of-distribution classes on each near-OOD and far-OOD benchmark. In addition, 20\% of the data is held out for unseen testing. The task for each category was randomly chosen from Table 1. In table 2, we describe the AUROC and average precision scores computed for Far-OOD detection between "Church (Outdoor)" and "Dining room" classes. Table 3 describes the AUROC and average precision scores computed for Near-OOD detection between "Living Room" and "Dining room" classes. For both these tasks, logistic regression and gradient boosting were used as the downstream model.

In the foundational implementation, the whole graph embedding model \cite{rozemberczki2020karate} is designed to assimilate visual concepts and features by integrating data from both in-distribution and out-of-distribution sources. The model learns to represent various characteristics by exploiting the relationships and similarities between different classes and domains. In a subsequent variation, the focus shifted towards exploring the model's zero-shot performance. In this experiment, the graph embedding model is trained exclusively on in-distribution data. After the training phase, the model's adaptability to out-of-distribution samples is assessed. This is executed by employing a one-class Support Vector Machine (SVM) with a Radial Basis Function (RBF) kernel. The one-class SVM can be formally represented as: $\text{minimize} \; \frac{1}{2}\|\mathbf{w}\|^2 - \rho$ subject to $\langle \mathbf{w}, \Phi(x_i) \rangle - \rho \leq \varepsilon$ for $i = 1, \ldots, n$ and $\rho \leq 0$. Here, \( \mathbf{w} \) denotes the normal vector to the hyperplane, while \( \Phi(x_i) \) represents the mapping of data points using the RBF kernel. The variable \( \rho \) is associated with the margin, and \( \varepsilon \) acts as a slack variable that permits certain misclassifications. This adaptation allows for an exploration of the model's potential to generalize across unseen datasets, shedding light on its capacities for zero-shot learning.

\cite{lee2018simple} proposed using the Mahalanobis distance for OOD detection by fitting a Gaussian distribution to the class-conditional embeddings. Conventionally, they let \(f(x)\) denote the embedding of an input \(x\) (e.g., the penultimate layer before computing the logits). We, instead, fit a Gaussian distribution to the whole graph embeddings, computing the per-class mean $\mu_c = \frac{1}{N_c} \sum_{i: y_i = c} f(x_i)$ and a shared covariance matrix: $\Sigma = \frac{1}{N} \sum_{c=1}^{K} \sum_{i: y_i = c} (f(x_i) - \mu_c) (f(x_i) - \mu_c)^T$. The Mahalanobis score (negative of the distance) is then computed as $\text{score}_{\text{Maha}}(x) = - \min_c \frac{1}{2} (f(x) - \mu_c)^T \Sigma^{-1} (f(x) - \mu_c)$. The performance of a downstream model is computed and presented.

\textbf{A note on explainability :} In our methodology, we represent images as visual concept networks where nodes are objects and edges depict their relationships. These graphs, uniform in node count (linked to the vocabulary) are distinct in topology, capture each image's unique configuration. Our analysis begins by encoding an image into a visual concept network, then projecting it onto a latent space Z, preserving key features. We then identify the nearest in-distribution graph to our target image's graph within this latent space. The crux of our explainability lies in the analysis of the topological differences between the target image's graph and its nearest in-distribution counterpart. By examining the disparities in node connections, we can uncover "why" an image is out of distribution.

\begin{figure*}[!ht]
    \centering
    \begin{subfigure}{0.32\textwidth}
        \centering
        \includegraphics[width=\linewidth]{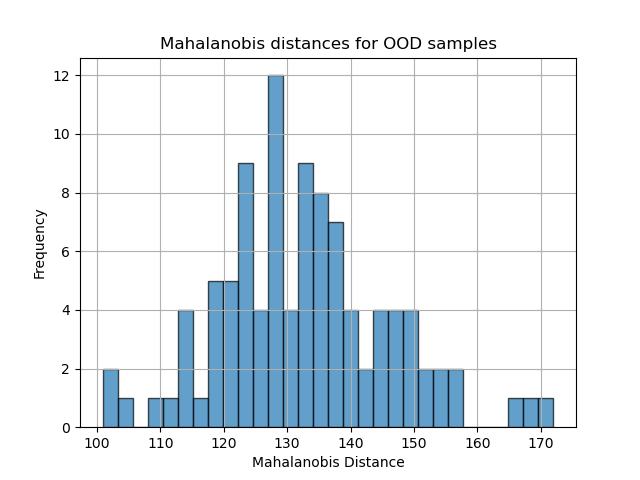}
        \caption{Mahalanobis Distance}
    \end{subfigure}%
    \hfill
    \begin{subfigure}{0.32\textwidth}
        \centering
        \includegraphics[width=\linewidth]{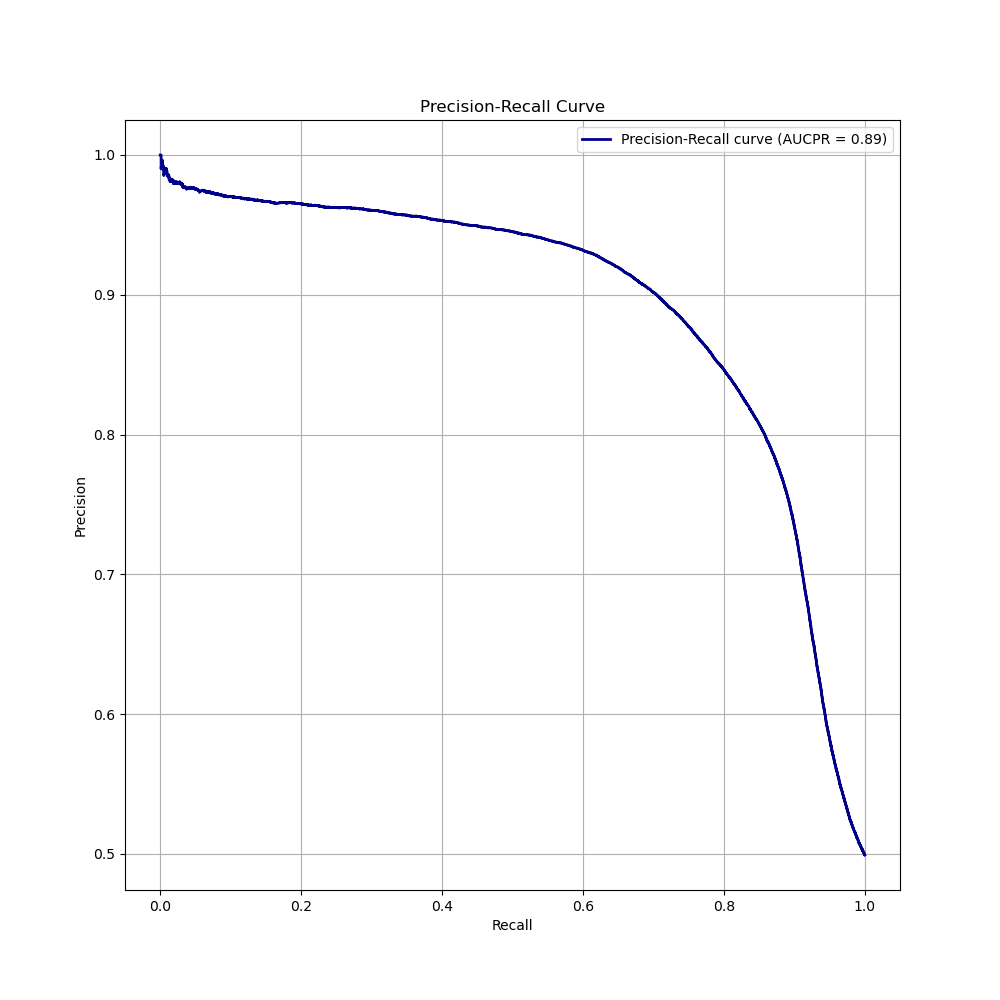}
        \caption{Precision-Recall Curve}
    \end{subfigure}%
    \hfill
    \begin{subfigure}{0.32\textwidth}
        \centering
        \includegraphics[width=\linewidth]{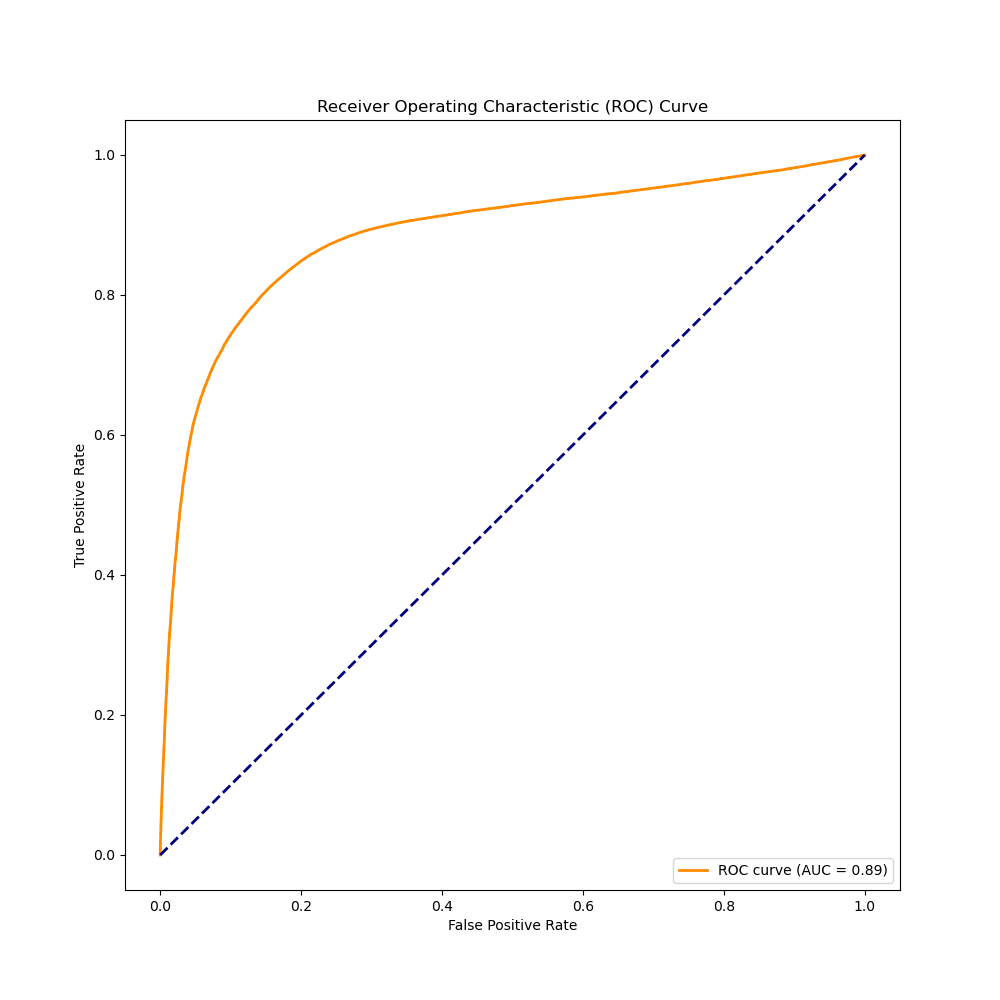}
        \caption{ROC AUC}
    \end{subfigure}
    \caption{Far OOD evaluation metrics while using class conditioned mahalanobis score. (a) Mahalanobis distance. (b) Precision-Recall Curve. (c) ROC AUC.}
    \label{fig:far_OOD_evaluation}
\end{figure*}

\begin{table*}[!ht]
\centering
\caption{Comparative evaluation of different graph embedding methods on the Far OOD task, using the COCO Vocabulary as a primary reference. Metrics include AUROC, AUPR, and F1-Score for both training and testing sets. Results are shown for Logistic Regression and Gradient Boosting classifiers. Additional results are provided for ablation studies using different vocabularies: Objects365, LVIS, and OpenImages.}
    \resizebox{\textwidth}{!}{

\begin{tabular}{lcccccccccccc}
\toprule
& \multicolumn{4}{c}{AUROC} & \multicolumn{4}{c}{AUPR} & \multicolumn{4}{c}{F1-Score} \\
\cmidrule(lr){2-5} \cmidrule(lr){6-9} \cmidrule(lr){10-13}
& \multicolumn{2}{c}{Logistic Reg} & \multicolumn{2}{c}{Grad Boost} & \multicolumn{2}{c}{Logistic Reg} & \multicolumn{2}{c}{Grad Boost} & \multicolumn{2}{c}{Logistic Reg} & \multicolumn{2}{c}{Grad Boost} \\
\cmidrule(lr){2-3} \cmidrule(lr){4-5} \cmidrule(lr){6-7} \cmidrule(lr){8-9} \cmidrule(lr){10-11} \cmidrule(lr){12-13}
& Train & Test & Train & Test & Train & Test & Train & Test & Train & Test & Train & Test \\
\midrule
Graph2Vec & 89.91 & 88.84 & 90.74 & 88.65 & 85.59 & 85.8 & 86.5 & 86.1 & 89.98 & 88.47 & 90.85 & 88.07 \\
GL2Vec & 89.54 & 88.88 & 90.04 & 87.06 & 85.12 & 83.33 & 85.7 & 80.25 & 89.61 & 89.48 & 90.13 & 88.29 \\
FGSD & 90.15 & 86.02 & \textbf{91.50} & \textbf{91.46} & 86.02 & 86.36 & 89.29 & 89.37 & 86.35 & 89.29 & 91.17 & 91.16 \\
SF & 90.15 & 90.27 & 90.15 & 90.27 & 86.02 & 86.35 & 86.02 & 86.35 & 90.18 & 90.35 & 90.18 & 90.35 \\
NetLSD & 90.15 & 90.27 & 90.15 & 90.27 & 86.02 & 86.35 & 86.02 & 86.35 & 90.18 & 90.35 & 90.18 & 90.35 \\
FeatherGraph & 90.15 & 90.27 & 90.15 & 90.27 & 86.02 & 86.35 & 86.02 & 86.35 & 90.18 & 90.35 & 90.18 & 90.35 \\
LDP & 90.15 & 90.27 & 90.15 & 90.27 & 86.02 & 86.35 & 86.02 & 86.35 & 90.18 & 90.35 & 90.18 & 90.35 \\
Wavelet Char & 90.15 & 90.27 & 90.15 & 90.27 & 86.02 & 86.35 & 86.02 & 86.35 & 90.18 & 90.35 & 90.18 & 90.35 \\
\midrule 
COCO (vocab)& 89.91	& 88.84	& 90.74& 	88.65	& 85.59	& 85.80	& 86.50	& 86.10	& 89.98	& 88.47	& 90.85& 	88.07 \\
Objects365 (vocab)&  86.35	& 84.30	& 90.23	& 89.62 & 81.35	& 81.96	& 85.77	& 85.43	& 86.32	& 82.6	& 90.37	& 89.75 \\
LVIS	(vocab)& 85.64	& 86.17	& 88.08	& 88.42	& 79.89	& 80.94	& 82.76 & 83.47	& 85.97	& 86.45	& 88.38	& 88.76 \\
OpenImages	(vocab)& 85.61	& 83.9	& 88.58& 	88.39& 	79.88& 	79.96& 	83.16& 	83.11 & 	85.92 & 83.05 & 88.95 & 88.86 \\
\bottomrule
\end{tabular}
}
\end{table*}

\begin{table*}[!ht]
\centering
\caption{Comparative evaluation of different graph embedding methods on the Near OOD task, using the COCO Vocabulary as a primary reference. Additional results are provided for ablation studies using different vocabularies: Objects365, LVIS, and OpenImages.}
\label{tab:results_near_ood}
    \resizebox{\textwidth}{!}{

\begin{tabular}{lcccccccccccc}
\toprule
& \multicolumn{4}{c}{AUROC} & \multicolumn{4}{c}{AUPR} & \multicolumn{4}{c}{F1-Score} \\
\cmidrule(lr){2-5} \cmidrule(lr){6-9} \cmidrule(lr){10-13}
& \multicolumn{2}{c}{Logistic Reg} & \multicolumn{2}{c}{Grad Boost} & \multicolumn{2}{c}{Logistic Reg} & \multicolumn{2}{c}{Grad Boost} & \multicolumn{2}{c}{Logistic Reg} & \multicolumn{2}{c}{Grad Boost} \\
\cmidrule(lr){2-3} \cmidrule(lr){4-5} \cmidrule(lr){6-7} \cmidrule(lr){8-9} \cmidrule(lr){10-11} \cmidrule(lr){12-13}
Method & Train & Test & Train & Test & Train & Test & Train & Test & Train & Test & Train & Test \\
\midrule
Graph2Vec & 54.59 & 55.40 & 56.55 & 55.58 & 52.67 & 53.22 & 52.52 & 53.31 & 55.54 & 60.75 & 62.55 & 62.21 \\
GL2Vec & 55.23 & 53.56 & 56.64 & 54.35 & 52.79 & 52.14 & 53.58 & 52.60 & 57.50 & 65.44 & 62.60 & 61.45 \\
FGSD & 54.51 & 54.35 & \textbf{70.85} & \textbf{70.21} & 52.35 & 52.61 & 64.86 & 64.56 & 59.69 & 59.76 & 70.35 & 69.90 \\
SF & 56.07 & 55.86 & 56.07 & 55.86 & 53.23 & 53.47 & 52.23 & 52.47 & 62.73 & 62.75 & 62.73 & 62.75 \\
NetLSD & 54.95 & 52.90 & 56.08 & 55.86 & 52.68 & 52.90 & 53.23 & 53.47 & 51.85 & 51.82 & 62.74 & 62.75 \\
FeatherGraph & 56.07 & 55.86 & 56.07 & 55.86 & 53.23 & 53.47 & 53.23 & 53.47 & 62.73 & 62.75 & 62.73 & 62.75 \\
LDP & 54.95 & 54.72 & 56.07 & 55.86 & 52.68 & 52.90 & 53.23 & 53.47 & 51.85 & 51.82 & 62.73 & 62.75 \\
Wavelet Characteristic & 56.07 & 55.86 & 56.07 & 55.86 & 53.23 & 53.47 & 53.23 & 53.47 & 62.73 & 62.75 & 62.73 & 62.75 \\
\midrule
COCO (vocab) & 54.59 & 55.40 & 56.55 & 55.58 & 52.67 & 53.22 & 52.52 & 53.31 & 55.54 & 60.75 & 62.55 & 62.21 \\
Objects365 (vocab)& 58.53 & 58.44 & 59.28 & 58.93 & 54.98 & 55.18 & 55.56 & 55.69 & 56.71 & 59.7 & 56.45 & 55.67 \\
LVIS (vocab)& 60.32 & 59.75 & 60.54 & 60.09 & 56.3 & 56.07 & 56.49 & 56.49 & 57.78 & 60.62 & 57.58 & 57.59 \\
OpenImages (vocab)& 58.97 & 59.19 & 59.92 & 59.63 & 55.35 & 55.95 & 56.08 & 56.23 & 55.98 & 54.39 & 56.05 & 55.76 \\
\bottomrule
\end{tabular}
}
\end{table*}

\textbf{Task Characteristics: } Our findings show that Far OOD tasks are easier than Near OOD tasks due to the distinct visual differences between the categories. For example, an outdoor scene varies greatly from a living room, simplifying the Far OOD detection.

\textbf{Benchmarking Graph Embedding Techniques: } In benchmarking various graph embedding methods, we observed similar performances among some techniques and their associated models. This suggests these methods may extract similar graph features, leading to comparable decision-making patterns. Graph2Vec, GL2Vec, and FGSD stood out for their ability to capture structural graph details, particularly in Far OOD tasks. While some variations in performance were noted, the choice of downstream model was key to achieving the best results.

\textbf{Classifier Performance} The Gradient Boosting classifier consistently outperformed Logistic Regression in training and testing across most graph embeddings, notably in the AUROC metric. The close alignment between training and testing scores suggests minimal overfitting, highlighting the embeddings' effectiveness in both standard and OOD scenarios. This consistent performance across different embedders and classifiers reinforces the reliability of our graph representation methods for OOD detection.

\textbf{Ablation Studies with Richer Vocabularies :}  Our ablation study explored the impact of using extensive vocabularies, like COCO \cite{lin2014microsoft} with 80 concepts, Objects365 \cite{shao2019objects365} with 365, OpenImages \cite{kuznetsova2020open} with 500, and LVIS \cite{gupta2019lvis} with over 1200. While a broader vocabulary increases the model's expressive power, it also leads to larger, less interconnected visual concept networks, posing challenges in creating meaningful embeddings. Nonetheless, our results showed consistent performance across these vocabularies, demonstrating the robustness of our graph embedding techniques.

\begin{table}[!ht]
\centering
\caption{Comparison of different embedders on zero-shot Far-OOD and Near OOD tasks, using one-class SVM. }
\begin{tabular}{lcc}
\toprule
Embedder & AUROC & AUPR \\
\midrule
\multicolumn{3}{c}{\textbf{Far-OOD}} \\
FGSD & 0.5372 & 0.5372 \\
SF & 0.3851 & 0.3851 \\
NetLSD & 0.4920 & 0.4920 \\
FeatherGraph & \textbf{0.8526} & \textbf{0.8526} \\
LDP & 0.4841 & 0.4841 \\
WaveletCharacteristic & 0.8526 & 0.8526 \\
\midrule
\multicolumn{3}{c}{\textbf{Near OOD}} \\
FGSD & 0.5132 & 0.5132 \\
SF & 0.5192 & 0.5192 \\
NetLSD & 0.5391 & 0.5391 \\
FeatherGraph & 0.4847 & 0.4847 \\
LDP & 0.5154 & 0.5154 \\
WaveletCharacteristic & 0.4847 & 0.4847 \\
\bottomrule
\end{tabular}
\end{table}

\textbf{Zero-shot performance:} Using a one-class SVM, graph embedders showed varying efficacy in zero-shot Far-OOD and Near OOD tasks. For Far-OOD, \textit{FeatherGraph} and \textit{WaveletCharacteristic} excelled in identifying vastly different visual concepts, contrasting with the struggles of \textit{SF} and average results from \textit{FGSD}, \textit{NetLSD}, and \textit{LDP}. However, in Near OOD tasks, all models' performances converged, nearly resembling random guesses, highlighting the challenge of differentiating similar visual categories. Notably, \textit{NetLSD} showed a slight advantage, but former Far-OOD leaders like \textit{FeatherGraph} and \textit{WaveletCharacteristic} matched the performance of \textit{LDP} and \textit{SF}. This shift emphasizes the complex nature of Near OOD tasks and the need for graph embeddings specifically tailored to capture these subtle distinctions.

\begin{table}[!ht]
\centering
\caption{Performance of RandomForest (RF) and GradientBoosting (GB) classifiers using the Graph2Vec embedder on the ImageNet dataset (multiclass) distribution detection. Metrics are weighted averages}
\begin{tabular}{lcccc}
\toprule
Classifier (Set) & Accuracy & AUC & AUPR & F1-Score \\
\midrule
RF (Train) & 1.0 & \textbf{1.0} & 1.0 & 1.0 \\
RF (Test) & 0.6045 & 0.6612 & 0.5554 & 0.5807 \\
\midrule
GB (Train) & 0.6239 & \textbf{0.7135} & 0.6284 & 0.6054 \\
GB (Test) & 0.6089 & 0.6672 & 0.5638 & 0.5865 \\
\bottomrule
\end{tabular}
\end{table}

\textbf{ImageNet Performance: } Our ImageNet study, leveraging the WordNet hierarchy, aimed at evaluating category recognition. Limited by computational resources, we used the Graph2Vec embedder with RandomForest (RF) and GradientBoosting (GB) classifiers. RF, while showing perfect training scores (accuracy, AUC, AUPR, F1-Score all at 1.0), suggested overfitting due to its lower test performance. GB displayed more balanced training and testing results, indicating a better-generalized model. Overall, GB proved slightly more effective than RF in broader ImageNet category detection.

\textbf{Mahalanobis Score based performance :}  Using Lee et al.'s (2018) approach, Gaussian distributions were fitted to graph embeddings. The Mahalanobis distances for OOD samples showed a clear Gaussian pattern (Figure 2), validating its use for OOD detection in our study. The model's high accuracy is evidenced by an AUCPR of 0.89 and a matching ROC curve AUC.

\textbf{Constraints:} Our approach hinges on an open vocabulary object detector to pinpoint key features within a specified dataset and domain. Ideally, this detector should also recognize features likely in out-of-distribution data. Given a complete feature list, graphs in $\mathbb{G}$ would have nodes for each feature, with edges formed based on observed pairwise relationships. This method of utilizing auxiliary models for verifying decisions mirrors the essence of boosting, fitting successive models to prior residuals. Such a structure could potentially lower the chances of operational failures.

\section{Conclusion}
Our paper introduces the use of graph-based representations of visual semantics for more effective out-of-distribution (OOD) detection across two novel tasks. By transforming outputs from pre-trained object-detection networks into semantic graphs, we enhance the AI's ability to discern and rationalize OOD instances. This method not only boosts OOD detection accuracy but also offers clear explanations for AI decisions, crucial for building trust in AI applications.

We thoroughly compare different graph embedding algorithms within our framework, illustrating the strengths of graph representations in OOD detection and guiding the selection of the most suitable embedding techniques. Our findings highlight the potential of graph methods for improving both interpretability and efficacy in OOD detection. This paper charts a new course in OOD detection research, spotlighting the value of semantic graphs and graph-based learning, and is poised to influence future explorations in OOD detection and AI.

\bibliographystyle{splncs04}
\bibliography{references}
%
%
%
%




\end{document}